# Constructing a Hierarchical User Interest Structure based on User Profiles

Chao Zhao
School of Computer
Science and Technology
Harbin Institute of Technology
Harbin, China
Email: zhaochaocs@gmail.com

Min Zhao
Baidu Inc.
Beijing, China
Email: zhaomin@baidu.com

Yi Guan
School of Computer
Science and Technology
Harbin Institute of Technology
Harbin, China
Email: guanyi@hit.edu.cn

*Abstract*—The interests of individual Internet users fall into a hierarchical structure which is useful in regards to building personalized searches and recommendations. Most studies on this subject construct the interest hierarchy of a single person from the document perspective. In this study, we constructed the user interest hierarchy via user profiles. We organized 433,397 user interests, referred to here as "attentions", into a user attention network (UAN) from 200 million user profiles; we then applied the Louvain algorithm to detect hierarchical clusters in these attentions. Finally, a 26-level hierarchy with 34,676 clusters was obtained. We found that these attention clusters were aggregated according to certain topics as opposed to the hyponymy-relation based conceptual ontologies. The topics can be entities or concepts, and the relations were not restrained by hyponymy. The concept relativity encapsulated in the user's interest can be captured by labeling the attention clusters with corresponding concepts.

## I. INTRODUCTION

Designing systems to retrieve or recommend relevant information to meet users' needs is an increasingly challenging endeavor as the massive stores of online data continue to grow. The most common approach to doing so is based on personalized information related to user interests. These individual interests, which we refer to here as *attentions*[1], are stored in the user profiles. They are a key component in the filtering and recommendation systems of today's web services, e.g., search engines and feeds[1], [2].

User attentions can be built into a hierarchy, where the general and specific interests of users are effectively organized in different levels. According to [3], higher-level interest categories reflect longer-term user interests; lower-level categories reflect the user's current interests. [4] attributes higher-level attentions to implicit, passive interests, while lower-level attentions correspond to explicit, active interests. In any case, this kind of user profile can be effectually utilized for personalization (e.g., query disambiguation, interest expansion, and cold-start problems alleviation) by capturing the semantic or knowledge-level similarities of attentions [5], [6], [7], [8], [9].

Figure 1 shows an example of "Sports-Basketball-NBA-Rockets" as a certain path in the interest hierarchy. Upon capturing these concepts in the user profile, the web search engine is more likely to ascribe the query to the Houston Rockets basketball team rather than spacecraft or other technically unrelated topics when the user simply searches the word "rockets".

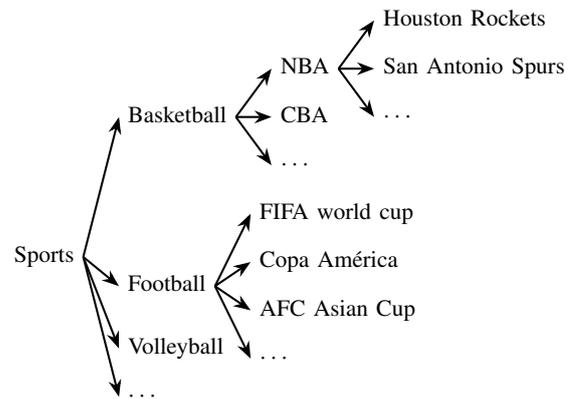

Fig. 1: Example interest hierarchy.

There are many existing methods for extracting hierarchical user interests from the user's behavior (e.g., clicks on or scrolls through various documents or sites). Previous studies have mainly focused on mining the interests of only one individual at a time; research on the inner structure or relativity among these interests themselves is relatively scant. For example, is this hierarchy similar to that of a conceptual ontology? What broader subjects will capture a user's interest, say, if he or she clicks on articles about artificial intelligence (AI)? How similar are this person's interests to another person who is interested in machine learning?

In this study, we attempted to construct an attention hierarchy directly from a set of user profiles, rather than a single profile, using a clustering technique. We explored the factors responsible for aggregating the attentions together into the hierarchy. Our primary contributions are two-fold:

- We organized all the attentions as a user attention network (UAN) based on their co-occurrences in user profiles,

---

[1]For clarity, we use *attentions* to refer to words representing user interests, e.g., "Harry Potter" or "Fantasy movie".

then divided the UAN into communities constituting an attention hierarchy structure.
- We analyzed the factors responsible for clustering certain attentions according to the attention hierarchy, and found that topics – not concepts – were paramount. Concepts only occasionally play a role similar to topics.

The remainder of this paper is structured as follows. In Section II, we give a brief review of related works on the hierarchical representation of user interests. In Section III, we describe the UAN construction and corresponding community detection method in detail. We present the community detection results in Section IV and discuss the aggregation mechanics of attentions in Section V. A brief conclusion and discussion on future research directions are presented in Section VI.

## II. RELATED WORKS

The extant research on constructing user interest hierarchies can be split into two main categories: Concept-based and content-based.

Concept-based methods model the hierarchy of interests with the aid of concept taxonomies. These include the Open Directory Project[10], [11], [6] or Wikipedia folksonomy[12], [13]. [10] used the first three levels of ODP categories to construct a general interest profile for mapping user queries to a set of categories. [11] mapped each interest of one user as a hierarchical ODP path, then calculated the similarity between the search results and the user profiles and re-ranked the results accordingly for the sake of personalization. [12] mined Twitter user's interests from the entities of their Tweets by finding corresponding concepts in Wikipedia folksonomy. By leveraging the Wikipedia category graph, [13] further constructed interest hierarchies of Twitter users via spreading activation.

Content-based methods involve constructing interest hierarchies without the aid of predefined taxonomy. [14], [15] mined concept-based user profiles from clickthrough data to infer user concept preferences, assuming that general terms with higher frequency would be placed at higher levels in the user profile hierarchy while specific terms with lower frequency were placed at lower levels. They defined two types of relationships, similarity and parent-child, for the concepts $c_1$ and $c_2$ in the search results and utilized the similarity measure $Sim(c_1, c_2)$ and conditional probability $P(c_1|c_2)$ and two thresholds to justify the existence of relationships between concepts. [4] extracted the terms from web pages bookmarked by users as their interests, and utilized a divisive clustering algorithm to group these interests into a hierarchy. The correlations between any two terms were measured by the Augmented Expected Mutual Information (AEMI) and the MaxChildren threshold-finding method was used to find a reasonable threshold for correlations. [16] identified mismatching between topic hierarchies used in analytics and learned from log data, which they resolved by first aggregating search queries and clicks into concepts in the hierarchy, then mapping the concepts to a topic taxonomy.

## III. METHODS

### A. User profile

We randomly selected 200 million user profiles from the Baidu search engine, which were obtained automatically from user behaviors (e.g., user search and click histories), to construct our dataset. Each user is associated with a series of attentions given personalized weights. The weight of each attention is an integer between 1 to 2,000 that indicates the importance of this attention to the user. Most of the attentions are entities, such as the "Harry Potter", "Stephen Curry", or "Fan Bingbing"; several were also be concepts, e.g., "fantasy novel", "NBA", or "film star". To improve the quality of the dataset, we only reserved the core attentions of each user with weights of exactly 2,000. We also discarded any users with less than 10 core attentions, because we prefer to believe that these attentions co-occur randomly rather than due to non-trivial factors, e.g., higher-level interests of users.

We obtained a total of 433,379 unique attentions after filtering. We analyzed the distribution of the number of attentions for an individual user as shown in Figure 2: One user can have no more than 250 core attentions. The distribution is power-law like, and most of the users have less than 50 attentions. We also identified the distribution of the number of followers for a unique attention, as shown in Figure 3. Both axes are logarithmic-scaled and explicitly follow a power-law distribution. There are a few attentions that possess a very large amount of followers, while most attentions are associated with relatively few people. In effect, attentions are very personalized attributes for individual users. The most popular concept and entity attentions are "society" and "Fan Bingbing", respectively.

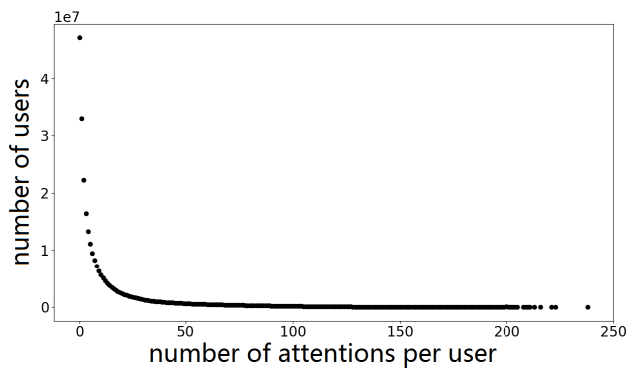

fig. 2: Distribution of number of attentions in an individual user profile. x-axis represents the number of attentions in the user profile; y-axis represents corresponding user numbers.

### B. User attention network

Attentions in one user profile can be used to construct the interest hierarchy for the current user, but are less helpful as far as determining the more general associations among attentions themselves. To capture these associations, we assumed that two attentions in one user profile are related: The more frequently the two attentions co-occur in one user profile, the more relevant the two attentions are.

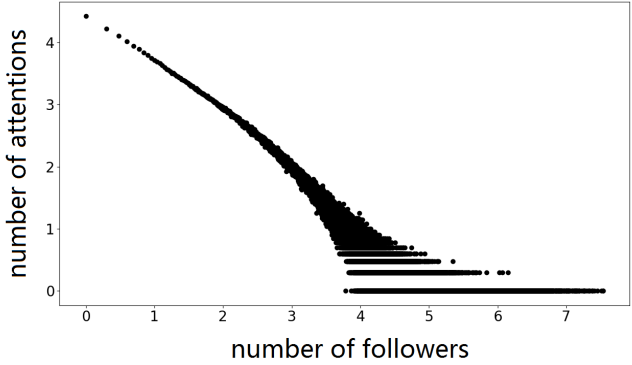

Fig. 3: The distribution of the number of users for a unique attention. X-axis represents the number of followers of an attention, and y-axis represents the corresponding attention numbers. Both two axes are logarithmic scaled.

Based on this assumption, we first constructed a complete, undirected graph for each user profile in which nodes are attentions and the weight of each edge is assigned as 1. We then merged all these graphs into a complex network called the user attention network (UAN). The weight of the same edge was cumulated during the merge process, and finally represented the number of users associated with both attentions. This weight represents the strength of association between the two attentions to some degree, but the association strength is not necessarily strictly proportional to the weight; it also depends on the user count of each attention. For example, there are 300 users with both attentions "J. k. Rowling" and "Harry Potter", but 20,000 users are interested in both "Harry Potter" and "society". In fact, the count of followers of "society" exceeded 3 million with 20,000 co-occurrences, and therefore is trivial in comparison. To get a more reasonable weight, attentions with large user count need to be penalized.

To ensure the edge weights of UAN can better depict the association strength of attentions, we re-calculated the weight of two attentions $a_i$ and $a_j$ as follows:

$$w(a_i, a_j) = \log(1 + w'_{ij}) \times \log(\frac{N}{UC_i + UC_j})$$

where $w'_{ij}$ is the original weight, and $UC_i$ and $UC_j$ are user counts of two attentions. $N$ is a large integer to make sure that the second term is positive, which we arbitrarily set as the number of users in our dataset. The first item of this formula decreases the impact of weight from linear to log-linear, while the second penalizes the weight according to the user count of two attentions.

Table I shows the five nearest attentions with "Harry Potter" before and after the weight re-calculation. The re-calculated results perform much better than the original weights.

TABLE I: The comparison five nearest neighbors of "Harry Potter" before and after weight re-calculation.

| before | | After | |
|---|---|---|---|
| neighbor | weight | neighbor | weight |
| society | 20052 | Daniel Radcliffe | 141.774 |
| Qiao Renliang | 12606 | Hermione Granger | 140.087 |
| Love O2O | 11898 | Emma Watson | 138.945 |
| entertainment | 11612 | Fantastic Beasts | 135.415 |
| doctor | 10892 | Harry Potter 5 | 134.320 |

### C. Attention community detection

From a complex network perspective, the UAN has a community-based structure. Like Figure 4 shows, several attentions in the UAN are connected more densely than others. These densely connected clusters (communities) of attentions implicitly indicate that they are more likely to appear simultaneously in a single user profile, from which the structure of the attention set can be inferred.

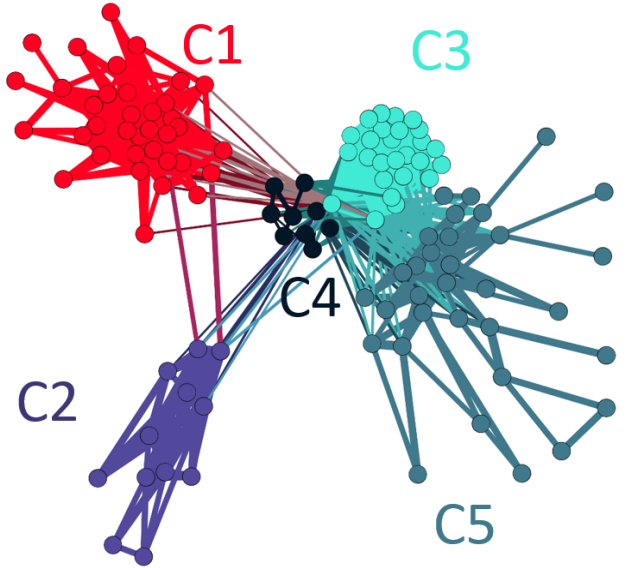

Fig. 4: An toy example of UAN.

The Louvain method (LM) [17] is one of the most widely used algorithm to discover hierarchical communities of nodes from large-scale weighted graphs. Compared with other community detecting algorithms, LM is time-efficient and easy-to-implement. Therefore, we apply it here to identify attention clusters in the UAN. It determines the community label of each node in an iteratively, greedy manner by maximizing a modularity measure $Q$[18]. Given a partition result, the modularity measures the difference between the edge density

in the inner community and that in a random network:

$$Q = \frac{1}{2m} \sum_{ij}(A_{ij} - \frac{k_i k_j}{2m})\delta(c_i, c_j)$$

where $c_i$ denotes the community of node $i$ and $\delta(i,j)$ is the Kronecker delta. $A$ is the adjacency matrix of the network. $m$ is the total number of edges. $\frac{1}{2}\sum_{ij} A_{ij}\delta(c_i,c_j)$ counts all the edges within the same community, and $\frac{1}{2}\sum_{ij}\frac{k_i k_j}{2m}\delta(c_i,c_j)$ is the expected number of edges of these communities in a random network. $Q$ is obtained by normalizing the difference of this two with the number of edges $m$. Higher modularity means that there are more edges in the inner of community than that expected by chance, and further indicates the existence of potential community structures in the current partition result.

LM first assigns individual communities to each node, then runs two phases iteratively to maximize the modularity. In the first phase, it repeatedly and sequentially merges each node $i$ to the community of one of its neighbors to ensure the maximum increase in modularity until modularity can no longer be improved by an individual move. Based on this result, the second phase regards the detected communities as new nodes to generate a new graph to which the first phase is reapplied. Once the maximum modularity is attained, the algorithm outputs a hierarchical clustering result.

Because the LM uses a heuristic method to determine the community assignment of each node, the result may not be optimal and the divided subgraphs may be still too large. To obtain smaller communities, we ran the Louvain method on the large results iteratively until two terminate conditions were reached:

- The current graph's node size is less than 1,000;
- The current graph's largest detected community covers more than 80% of the nodes.

The hierarchical structures of the results are reserved after each iteration, ultimately yielding an attention hierarchy in which all intermediate nodes and leaves can be regarded as attention clusters. (Leaves are the clusters with minimal sizes.)

### D. Community labeling

Attentions in the same communities of the hierarchy co-occur more frequently than those belonging to other communities. In other words, if the user expresses the interest within a certain community, he or she is likely to be interested in other attentions in the same community. By analyzing the common characteristics of attentions in the same hierarchy, we expected to find the underlying reasons that certain attentions were aggregated. It is cumbersome to analyze and summarize these patterns manually, however, since the counts of both attentions and communities are large.

To capture the more general and informative relations, we labeled each community with their corresponding concepts using the Baidu knowledge base (BKB), which provides the hypernym path of one attention. The BKB is a directed acyclic graph (DAG) and contains 13 domains (children) under the root node. For each domain, we used the lowest common ancestor (LCA) concept node of these attentions as labels to summarize the attentions in current community.

Cluster-based results inevitably contain noise, and the BKB may also not return the most appropriate fine-grained hypernym concepts. Both factors can lead to a strict LCA search returning coarse concepts. To avoid this, we relaxed the restrains to find only LCAs covering more than 50% of the attentions of the same domain in one community.

Figure 5 shows a mini version of the clusters containing seven attentions from two domains. The concept "war movie" contains 75% of the attentions in the *Culture* domain, which is marked with green blocks, while the concept "battle" contains all attentions in the *Life* domain marked in yellow. We labeled this cluster as "war movie & battle".

If more than one concept is labeled for a single community, we assume these two concepts have certain associations in regards to user interest level. For example, individuals who are interested in battles may also like films with battle scenes. Not only the leaf communities of the hierarchy, but all the intermediate communities can be effectively labeled with concepts. This can find the relations in a higher level, and avoid the ignorance of interesting concept relations due to the small size of clusters. Generally speaking, concept associations in smaller-size communities are stronger than those in the larger communities.

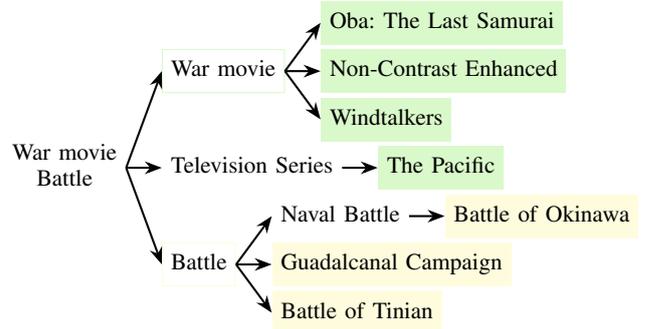

Fig. 5: Example of LCA labeling

## IV. RESULTS

After the community detection, we obtained an attention hierarchy which contained 34,676 small communities. Figure 6 shows the distribution of the number of communities and attentions in each level of the hierarchy. The shallowest leaf communities exist in the fourth level of the hierarchy, and the depth of the hierarchy is 26. Most of the communities and attentions are concentrated in the area from the 7th to the 19th level.

The distribution of community sizes is shown in Figure 7. Most of the communities have less then 30 attentions, and the largest size is no more than 1,000 under our partition strategy. Table II shows several communities from the attention hierarchy, which are just those have shown in Figure 4.

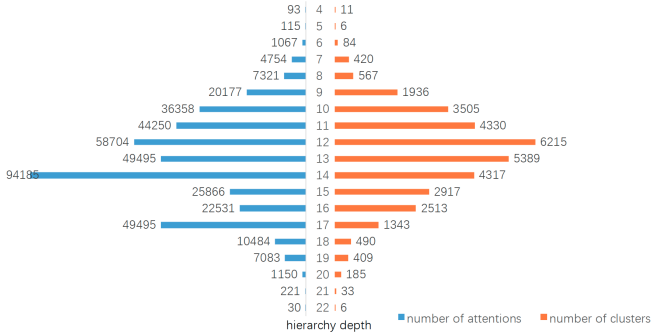

Fig. 6: Distribution of number of communities and attentions over the hierarchy depth.

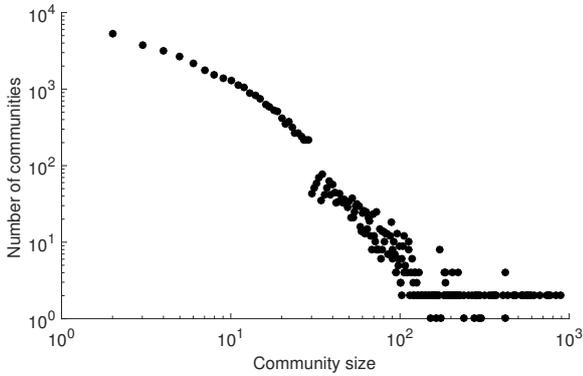

Fig. 7: Distribution of communities by size. X-axis represents the size of a single community; y-axis represents the corresponding community numbers. Both axes are logarithmic-scaled.

Unfortunately, it is difficult to measure the quality of our obtained communities. There are two kinds of indexes to evaluate a clustering result: internal or external. The former needs a pre-defined distance measure between two instances. Once we determine the weights of edges in UAN, the Louvian method can derive a reasonable clustering result accordingly. The latter index, as the name suggests, needs an external reference model (e.g., the clustering result given by experts) to compare with. However, there is no such a golden-standard model to tell us what the clusters of user attention should look like. We therefore have to evaluate the clusters manually, based on the common knowledge of humans. We randomly selected 200 communities from the result of LM, and then to label each as +1 (if more than half of the instances in the community belonged to the same topic or the relations among the topics are easily to understand) or -1 (other conditions). 84% of the sampled communities were labeled as positive, and indicated the reasonability of the our obtained communities to some degree.

TABLE II: Example clusters in table-structured hierarchy.

| | | |
|---|---|---|
| C1 | C1-1 | Sirius Black, Hedwig, Quirrell, Grindelwald, Hermione Granger, Luna, Snape, Moaning Myrtle |
| | | Quidditch, Ministry of magic, Felix Felicis, Patronus Charm, Expecto patronum, Broomsticks |
| | Ravenclaw, Hufflepuff, Gryffindor, sorting hat, Duemstrang Institute, Beauxbatons Academy of Magic, Butterbeer | |
| | Harry Potter, Harry Potter 5, Harry Potter cast, Daniel Radcliffe, Emma Watson, Bonnie Wright, Matthew Lewis, Michael Gambon, Alan Rickman, Tom Felton | |
| | house-elf, Voldemort cat, Dementor, Death Eater | |
| | James Porter, Viktor Krum, Alastor Moody, Weasley, Fleur delacour, Cedric Diggory, Neville Longbottom, Ribus Hagrid, Teddy Lupin, Frank Dillane | |
| C2 | Elijah Wood, Richard Armitage, Miranda Otto, Dominic Monaghan, Viggo Mortensen, Arwen | |
| | Elrond, Haldir, Earendil, Gil-galad, Glorfindel, Legolas Greenleaf | |
| C3 | artificial intelligence, cloud computing, silicon valley, YAHOO, unmanned driving, Internet + medicine, face recognition | |
| | Liu Zhen, Ren Zhengfei, Lei Jun, Jack Ma, Rothschild, Wanda, Rupert Hoogewerf | |
| | BAT, Alibaba stock, Google stock, Apple share price, Tencent market cap, Jingdong stock, Partner system, world's billionaires list | |
| | Internet e-commerce, Alipay, electronic business platform, WeChat payment, Micro payment, Maker, Self media | |
| C4 | Bodybuilding, McCarthyism, little pink, Zhonghua Jia, Magician, Round table, political correct, empathy, Q&A Platform, sinovation ventures, Carpe Diem | |
| C5 | machine learning, support vector machine, ID3, deep learning, decision tree | |

## V. DISCUSSION

Compared to the hierarchy structure of conceptual ontologies, in which concepts and entities are organized based on strict hyponymic relations, attentions in our hierarchy are organized by topic.

### A. The topic-based organizations

Attention organizations seem intuitively linked to hyponymic relations. For example, a person who is interested in the movie "Harry Potter" would more likely to be interested in other fantasy movies like "Lord of the Rings". According to our community result, however, none of other movies existed in the leaf community or siblings of "Harry Potter". These attentions instead centered around items like "Emma Watson", "Lord Voldemort", and "Quiddich". They are different concepts — actors, characters, and plot points of the film — but are organized under the same topic "Harry Potter". The community of "Lord of the Rings" also shows similar patterns comprised of actors in the film and its characters. "Harry Potter" or "Lord of the Rings" act as topics responsible for aggregating certain attentions together. They themselves are aggregated by more general topics, e.g., fantasy movies.

The biggest difference between topics and concepts is that concepts have strict top-down structures, are organized by single hyponymic relations, and follow the logical relation of transitivity. "Battle of Okinawa" is a naval battle and thus can falls under the larger "battle" category. The topic, however, is a less strict summary of a local attention set. It does not have to possess transitivity and the relations between attentions may vary. For example, "Emma Watson" falls under the topic "Harry Potter" but not "fantasy movie". The top-down structure is also not abstract. Sometimes a movie can be a topic and aggregate its cast members together, while actors can also form topics into which their films are aggregated together.

Further analysis of other communities indicates that such topic-based aggregation is common. A concept can sometimes play the role of a topic, but not always. Enriching the user's attention via general concepts is not always effective, as discussed above; further, enriching the user's attention to special cases at the concept level may also be ineffective. Consider again our questions posed at the start of Section I regarding interests shared by two users. If we know one person is interested in AI and another is interested in machine learning, the communities encapsulating these two attentions are as shown in C3 and C5 of Table II. The community containing "machine learning" falls under the topic "machine learning techniques", while the community containing AI and its near communities fall under the topic "Internet", which centers more around Internet companies as well as their founders, new technological products, and economic impacts.

These differences altogether demonstrate that the hierarchical structure of our user attentions is not well consistent with that of a conceptual ontology. Though "machine learning" is a hyponym of "AI" in conceptual ontology, most non-experts are disinterested in its concrete algorithms. On the other hand, "Harry Potter" and "Emma Watson", which fall into the same communities, have a low similarity in conceptual ontology as shown in Table III.

### B. Topic categories

There are a variety of entities or concepts that may play the role of topics. In the example "Harry Potter", the topic is a movie. The actors or even the director may also be the topic responsible for aggregating certain attentions related to the film together, however. These topics are mainly entities, and the corresponding relationships are the attributes of the topics. Concept-based topics, on the other hand, may aggregate their hyponyms as well as other concepts sharing certain similarities, like the "war film" and "battle" example described above.

TABLE III: Hypernym paths of "Emma Watson" and "Harry Potter".

| attention | Hypernym path |
| --- | --- |
| Emma Wason | Emma Watson → actor → entertainer → person |
| Harry Potter | Harry Potter → Fantasy film → film → film and television work → video work → work |

Another interest community is shown in C4 of Table II in which there do not seem to be any strong relationships among entities or concepts. The longtime users of Zhihu[2], however, would notice immediately that this cluster contains hot topics in the Zhihu community. In this case, issues of interest within a certain web community play the role of the topic.

The factors that define an entity or concept as the "topic" are still unknown. The category of a given concept itself is apparently not a factor. Though "Harry Potter" and "Oba: the Last Samurai" are both movies, for example, they play different roles in their respective communities.

### C. Adding topic relationships to conceptual ontology

Although the conceptual ontologies are not completely consistent with the interest hierarchies, they are still valuable resources in determining the similarities of two entities at the semantic or knowledge level. [3] used spreading activation to incrementally update the interest scores of concepts in user profiles, and then to calculate the similarities between user profile pairs based on the Euclidean distance of concept vectors. [19] applied the random walk algorithm to expand user profiles on the Wikipedia graph. They attempted to use inner Wikipedia links to alleviate the inconsistency between taxonomy and interest hierarchy, but inner links, arguably, fail to reflect entity relations across the user interest perspective.

To bring entities in the same topic closer together in the concept hierarchy, we tried to introduce the "related_to" relationships into the concept hierarchy based on the concept labels discussed in Section III-D. As shown in Figure 5, by adding "related_to" between the concepts "war movie" and "battle", we were able to reflect the relativity of these two concepts at the user interest level. Some other examples are shown at Table IV. Many network-structure based methods (e.g., spreading activation, random walking) can benefit from introducing new relations during the similarity calculation.

## VI. CONCLUSION

In this study, we explored the organization structure of user attentions (interests) from user profiles aggregated via a clustering method. We linked all attentions in user profiles into a user attention network (UAN), and utilized the Louvain method to obtain the hierarchical attention clusters from the UAN. We found that the user attention hierarchy is mainly organized by the topics, as opposed to hyponymy-relation-based conceptual hierarchies. Topics can be entities or concepts. The factors responsible for aggregation can be attribute relations, hyponymy relations, or more general similarities. Through the clustering results of user attentions, we were able to supply "related_to" relations to the conceptual ontologies to better depict the concept relativities across the user interest perspective.

---
[2]A Quora-like QA website in China

TABLE IV: Example concepts labeled to the same cluster.

| |
|---|
| war, war movie, troop, empire, militarist, tactic, arm |
| athlete, sport competition, sport award, competition schedule, sports channel, live platform, sport lottery |
| university, major, university rankings, recruitment, recruitment website |
| plane ticket, flight, hotel, travel agency, visa, tariff |
| estate, house price, building materials, building design, second-hand house |

In the future, we plan to explore the factors that define an entity or concept as a topic. We also will experimentally demonstrate the benefit of user attention hierarchy for personalized searches and recommendations.